\documentclass[10pt,twocolumn,letterpaper]{article}

 \usepackage[pagenumbers]{cvpr} 

\usepackage{threeparttable}
\usepackage{multirow}
\usepackage{verbatim}
\usepackage[table]{xcolor}
\usepackage{colortbl}

\definecolor{cvprblue}{rgb}{0.21,0.49,0.74}
\usepackage[pagebackref,breaklinks,colorlinks,citecolor=cvprblue]{hyperref}

\title{Segment Together: A Versatile Paradigm for Semi-Supervised Medical Image Segmentation}

\author{Qingjie Zeng$^{1^*}$~~~Yutong Xie$^{2^*}$~~~Zilin Lu$^{1}$~~~Mengkang Lu$^{1}$~~~Yicheng Wu$^{3}$~~~Yong Xia$^{1^\dagger}$\\
$^{1}$ School of Computer Science and Engineering, Northwestern Polytechnical University, China\\
$^{2}$ Australian Institute for Machine Learning, University of Adelaide, Australia\\
$^{3}$ Department of Data Science \& AI, Faculty of Information Technology, Monash University, Australia\\
{\tt\small \{maxwell,luzl,lmk\}@mail.nwpu.edu.cn,} \\
{\tt\small yutong.xie678@gmail.com, yicheng.wu@monash.edu, yxia@nwpu.edu.cn}
}

\begin{document}
\maketitle

\begin{abstract}
Annotation scarcity has become a major obstacle for training powerful deep-learning models for medical image segmentation, restricting their deployment in clinical scenarios. To address it, semi-supervised learning by exploiting abundant unlabeled data is highly desirable to boost the model training. 
However, most existing works still focus on limited medical tasks and underestimate the potential of learning across diverse tasks and multiple datasets. 
Therefore, in this paper, we introduce a \textbf{Ver}satile \textbf{Semi}-supervised framework (VerSemi) to point out a new perspective that integrates various tasks into a unified model with a broad label space, to exploit more unlabeled data for semi-supervised medical image segmentation.
Specifically, we introduce a dynamic task-prompted design to segment various targets from different datasets. Next, this unified model is used to identify the foreground regions from all labeled data, to capture cross-dataset semantics. 
Particularly, we create a synthetic task with a cutmix strategy to augment foreground targets within the expanded label space. 
To effectively utilize unlabeled data, we introduce a consistency constraint. This involves aligning aggregated predictions from various tasks with those from the synthetic task, further guiding the model in accurately segmenting foreground regions during training.
We evaluated our VerSemi model on four public benchmarking datasets. Extensive experiments demonstrated that VerSemi can consistently outperform the second-best method by a large margin (e.g., an average 2.69\% Dice gain on four datasets), setting new SOTA performance for semi-supervised medical image segmentation. The code will be released.

\end{abstract}

\section{Introduction}
\label{sec:intro}
Medical image segmentation is a long-standing challenge~\cite{shen2017deep, cheplygina2019not, jiao2022learning, zeng2023dm}. Due to the scarcity of voxel-level annotations, semi-supervised learning (SSL), which can learn from limited labeled and abundant unlabeled data, has been applied to medical image segmentation tasks.

\begin{figure}[!t]
    \centering
    \includegraphics[width=0.46\textwidth]{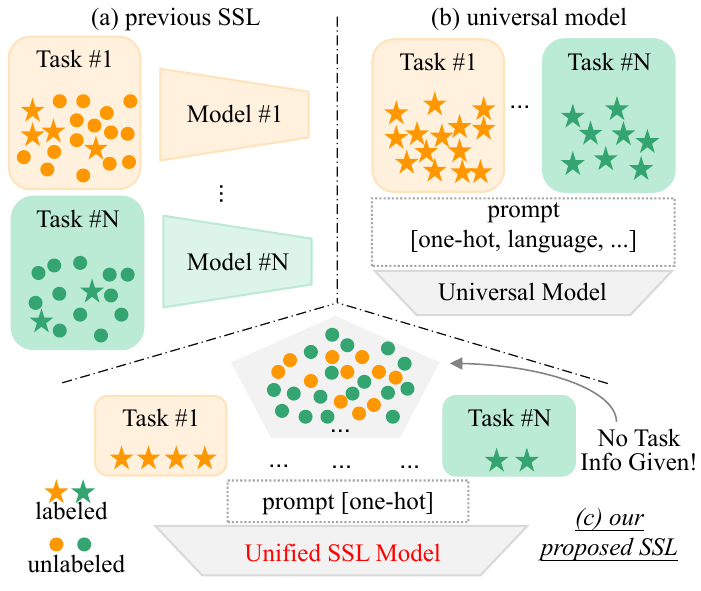}
    \caption{A brief illustration of the difference among previous SSL (a), universal model (b) and our proposed SSL (c). \textbf{First}, the previous SSL learns each model in isolation and neglects the importance of data integration. \textbf{Second}, universal models, $e.g.,$ DoDnet~\cite{zhang2021dodnet}, leverage diverse task prompts to acquire knowledge from multiple tasks in a supervised manner, which lacks the ability to handle unlabeled data when task info is unknown. 
    By comparison, our proposed SSL can not only complete various missions simultaneously but also learn from unlabeled data without requiring associated task info.}
    \label{illu}
\end{figure}

Generally, there are two popular SSL paradigms, \textit{i.e.}, pseudo-labeling~\cite{zeng2023pefat} and consistency regularization~\cite{van2020survey, wu2022exploring}. The former aims to find trustworthy pseudo-labels for re-training, $e.g.$, setting an adaptive threshold during the learning process to filter out unreliable predictions~\cite{zhang2022boostmis}. The latter focuses on making consistent predictions for different augmentations of the same input~\cite{sohn2020fixmatch, miyato2018virtual}. Despite their prevalence, most SSL methods are restricted to a specific task, where labeled and unlabeled data share an identical label space. 
Due to the insufficient supervision of the single task, the learned distributions of labeled and unlabeled data are prone to be inconsistent, resulting in poor generalizability and limited performance of SSL approaches. 
{Besides, the task-specific scenario often means that a significant portion of the unlabeled data, which might not fit perfectly into the predefined task label space, remains underutilized.}

Recently, universal models have drawn increasing research attention. They are trained on multi-domain and/or multi-modality data for multi-task in two ways.
The first approach involves pre-training the model using task-agnostic unlabeled data through self-supervised learning, followed by fine-tuning on task-specific data for individual downstream tasks~\citep{wang2023foundation, xie2022unimiss, zhou2022advancing}.
The second method trains a model jointly using multiple task-specific data in a supervised fashion~\citep{chen2023towards, ye2023uniseg, zhang2021dodnet}. 
Universal models have unveiled superior performance over traditional task-specific models on a variety of tasks, spanning both the computer vision~\cite{kirillov2023segment, xue2023ulip} and medical imaging communities~\cite{liu2023clip, ye2023uniseg, xie2022unimiss}. 
The success of universal models emphasizes the importance of integrating data and tasks to improve representation learning. 
This observation inspires us to amalgamate multiple datasets and tasks within the framework of SSL. Such integration promises not only to harness an expanded corpus of {unannotated and} annotated data, thereby bolstering the supervised component of SSL, but also to substantially enhance the model's generalization capabilities, thereby extending its applicability across diverse domains.

\begin{figure}[t]
    \centering
    \includegraphics[width=0.5\textwidth]{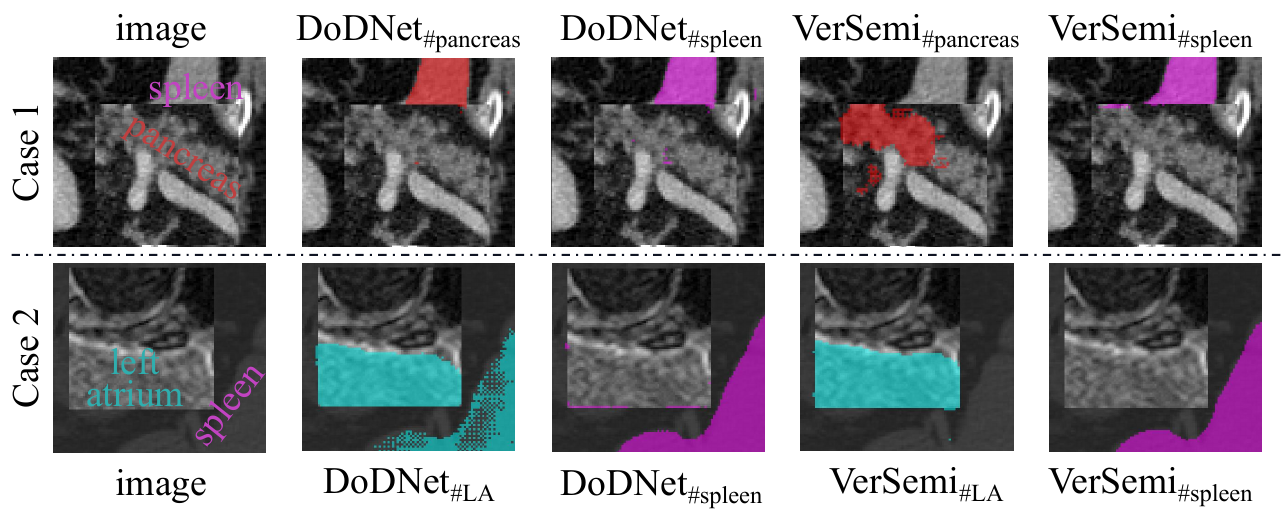}
    \caption{The prompt-weakening phenomenon. In \textbf{Case 1}, DoDNet fails to predict the region of pancreas, and can only recognize spleen voxels no matter under the prompt of pancreas or spleen. In \textbf{Case 2}, DoDNet mistakenly highlights the region of spleen when prompted by left atrium. By comparison, VerSemi has addressed this issue by devising an auxiliary constraint. (see Section~\ref{sec3.2})}
    \label{dod_uni}
\end{figure}

{In this paper, we introduce a \textbf{Ver}satile \textbf{Semi}-supervised framework (VerSemi), a novel approach that revolutionizes common SSL paradigms. 
\textbf{Firstly}, VerSemi surpasses task-specific learning constraints by integrating multiple targets into a unified framework. It seamlessly establishes an enhanced label space by amalgamating pertinent task labels, and accomplishes multiple tasks simultaneously with the assistance of a task-prompted dynamic head.}
\textbf{Secondly}, considering that task specifics are required for prompt-driven models to generate prompts~(one-hot encoding, language description, $etc$) during the learning process, an issue is raised that unlabeled data may not be mined if associated task information is remained unknown~(see Fig.~\ref{illu}).
To tackle this problem, VerSemi first constructs a synthetic task by leveraging cutmix on labeled data. In this way, the data in the synthetic task span a diverse range of foreground targets within the expanded label space.
By joint training with the synthetic task, VerSemi can recognize and segment all potential foreground regions. 
%
Grounded on this ability, VerSemi simplifies learning from unlabeled data by eliminating the need for task-specific details. This is achieved by ensuring consistency between combined predictions from relevant tasks and synthetic ones.
\textbf{Thirdly}, we empirically observe that prompts oftentimes do not work well,~\eg, models fail to recognize the object indicted by a specific prompt~(see Fig.\ref{dod_uni}). To address this issue, an auxiliary constraint is designed to regularize VerSemi to enhance its controllability when meeting task-specific prompts.
In addition, it is worth noting that current SSL methods can not directly realize task-agnostic unlabeled data learning, as they either demand a teacher model or extra sub-networks for supervision. Our contributions are three-fold.

\begin{itemize}
\item Different from current SSL methods that learn tasks individually, the proposed VerSemi performs well with the new setting of integrating various pertinent SSL tasks into a unified framework.
\item We achieve task-agnostic unlabeled data learning by devising a "synthetic task”. This design facilitates the learning of unified foreground segmentation. With this segmentation ability as a constraint, unlabeled data can be excavated without acquiring associated task specifics.
\item Extensive experiments on four public datasets validate the superiority of VerSemi, which presents remarkable improvements compared to task-isolated SSL models (\eg, BCP, CauSSL) and associated task-unified models (\eg, Uni-BCP, Uni-CauSSL).
\end{itemize}

\section{Related Work}
\label{sec:related}

\begin{figure*}[!t]
    \centering
    \includegraphics[width=0.9\textwidth]{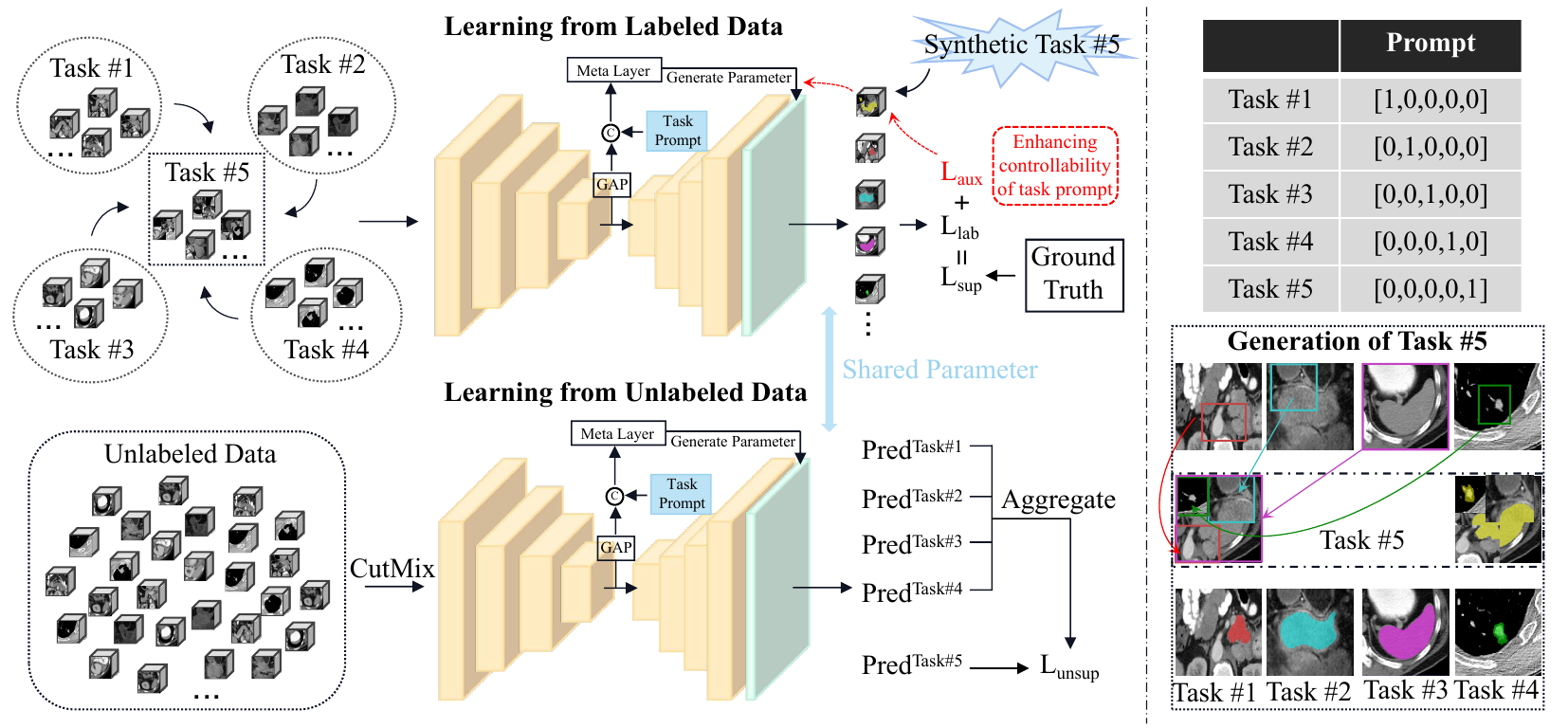}
    \caption{Illustration of VerSemi. VerSemi has a task-prompted dynamic head which can flexibly process different tasks at the same time, along that an auxiliary constraint $\mathcal{L}_{axu}$ is designed to augment the reliability of associated task prompt. During labeled data learning, we construct an synthetic task (Task\#5), which aims to segment all the foreground regions. As for unlabeled learning, the aggregated prediction prompted by Task\#1 $\sim$ Task\#4 is forced to be consistent with the prediction prompted by Task\#5, when feeding mixed unlabeled data into the model. Therefore, the proposed VerSemi does not require task information to learn from unlabeled data and is more versatile.}
    \label{framework}
    \vspace{-3mm}
\end{figure*}

\subsection{Semi-supervised Learning}
Semi-supervised learning (SSL)~\cite{chen2022semi, mey2022improved, cheplygina2019not, wu2022mutual} is emerged to mitigate the issue of tedious data annotation, by learning from unlabeled data with scarce labeled data. Many efforts are made to explore how to excavate information from unlabeled data adequately. For instance, UPS~\cite{rizve2020defense} reduced unreliable pseudo-labels by calibrating models with uncertainty. PEFAT~\cite{zeng2023pefat} investigated the probability distribution of pseudo-labeled data, and further proposed a selection standard from the perspective of loss distribution. SoftMatch~\cite{chen2022softmatch} and FreeMatch~\cite{wang2022freematch} tried to address the quantity-quality trade-off issue with adaptive threshold. However, these SSL frameworks focus on learning labeled and unlabeled data within each single task, followed by an issue of whether they have the ability to scale up to heterogeneous tasks. Beyond that, another problem is raised up when learning tasks individually, that is the improvement is constantly marginal due to insufficient representation acquired from limited labels in each dataset. To address the mentioned issues, we advocate learning a unified SSL model with an integrated dataset, under the new setting of learning various pertinent SSL tasks concurrently.

\subsection{Representation Learning with Integrated Data}
To improve the model performance and representation ability, some works propose to learn a unified model that can complete multiple tasks simultaneously, rather than training task-specific model separately~\cite{kim2022universal, lee2023unified}. For example, DoDNet~\cite{zhang2021dodnet} collected an abdominal dataset from seven partially labeled datasets for model training, and presented better-averaged results than training on every single dataset. CLIP-Driven Universal Model~\cite{liu2023clip} further advanced this idea by introducing CLIP embedding~\cite{radford2021learning} to help the model capture anatomical relations between different tumors and organs. UniSeg~\cite{ye2023uniseg} leveraged different modalities including CT, MRI and PET, whose performance surpassed those models trained with a single modality. 
These studies underscore the importance of robust data engines, emphasizing the need to leverage as much accessible data as possible. While the majority of these investigations focus on either fully supervised learning~\cite{ye2023uniseg, liu2023clip} or self-supervised learning~\cite{xie2022unimiss}, rare of them utilize simultaneously labeled and unlabeled data collected from different tasks. By contrast, we propose VerSemi and make an exploratory attempt.

Additionally, it is necessary to mention that compared to DoDNet~\cite{zhang2021dodnet}, a method leverages one-hot task prompt to learn from different tasks, our proposed VerSemi differs in: (1) DoDNet is designed under fully-supervised setting, which can not handle unlabeled data if the associated task information are not given; and (2) there exists severe prompt-weakening phenomenon during task learning procedure~(see Fig.~\ref{dod_uni}), this situation is overlooked by prompt-driven model like DoDNet. And VerSemi tackles this issue by designing an auxiliary constraint.

\section{Method}


Fig.~\ref{framework} shows the pipeline of our proposed VerSemi model, integrating various semi-supervised segmentation tasks from different datasets into a unified framework. 
Here, Section~\ref{sec3.1} shows the dynamic kernel generation in our VerSemi. Then, Sections~\ref{sec3.2} and \ref{sec3.3} further delve into the details of exploiting limited task-aware labeled data and task-agnostic unlabeled data for training, respectively.

\subsection{Dynamic Convolution with Task Prompt}
\label{sec3.1}

Despite the great progress of deep-learning based works, it remains sub-optimal for an individual model with fixed convolutional kernels to handle different segmentation tasks at the same time~\cite{zhang2021dodnet, lei2022semi}. To improve the performance, a common practice is to use a multi-head architecture, but it suffers from severe computational overhead with the increase of on-coming tasks, thus is not suitable when confronted with multiple tasks. To reduce the computational burdens of different heads, we utilize dynamic filter generation to construct the segmentation head, which can adaptively process different tasks with task-specific prompts without extra costs. The filter generation is defined as:
\begin{equation}
\begin{aligned}
    w_{k} &= \psi(GAP(Embedding), [Prompt_{\#k}];\theta_{\psi}) \\
      \mathcal{P}_{k} &= SoftMax({f}_{D}(Embedding)*w_{k}),
  \label{eq:one}
\end{aligned}
\end{equation}
where $\psi$ is one convolutional layer with parameter $\theta_{\psi}$, which is employed to dynamically generate parameters $w_{k}$ for the current Task\#k with $[Prompt_{\#k}]$. Here we set the prompt in a one-hot encoding format, which is then concatenated with global averaged feature embedding before feeding into $\psi_{\theta}$. $\mathcal{P}_{k}$ is the prediction for Task\#k, $f_{D}$ is the decoder and symbol $\ast$ represents convolution. With $[Prompt_{\#k}]$, VerSemi can accurately perceive the ongoing task and flexibly adapt kernels to fit it.

\begin{figure}[t]
    \centering
    \includegraphics[width=0.48\textwidth]{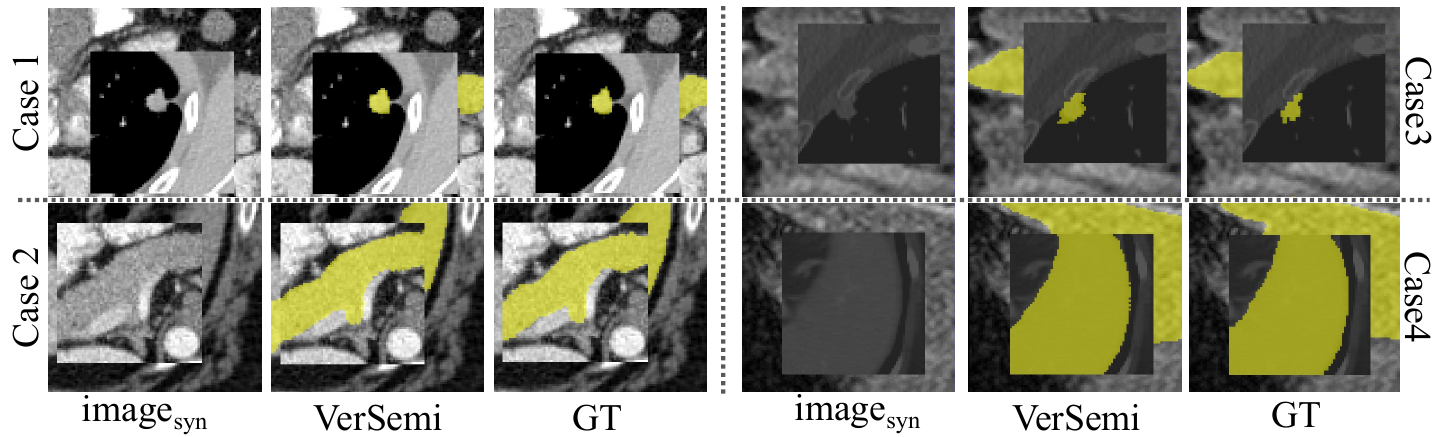}
    \caption{Predictions made by VerSemi when facing synthetic data. \textbf{Case 1} and \textbf{Case 3}: cutmix between left atrium and lung tumor. \textbf{Case 2}: cutmix between pancreas and spleen.
    \textbf{Case 4}: cutmix between spleen and left atrium.We can find VerSemi can produce accurate masks if prompted by Task\#5.}
    \label{cutmix_vis}
    \vspace{-3mm}
\end{figure}

\subsection{Task-aware Labeled Data Learning}
\label{sec3.2}
In this work, four pertinent tasks are incorporated. \textbf{Task\#1} $\sim$ \textbf{Task\#4} are the pancreas, left atrium, spleen and lung tumor segmentation tasks. Below, we describe the construction and utilization of the additionally synthesized \textbf{Task\#5}.

{\noindent\bf Generation of Task\#5.} As the bottom-right of Fig.~\ref{framework} shows, we construct a synthetic task (Task\#5) based on labeled data from pertinent tasks (Task\#1\ $\sim$ Task\#4). Task\#5 is built to help VerSemi achieve task-agnostic learning from unlabeled data and guide VerSemi to segment all foreground regions when facing mixed data~(see Fig.~\ref{cutmix_vis}). The data generation of Task\#5 is formulated as:
\begin{equation}
\begin{aligned}
  \mathcal{X}_{syn(i,j)}^{l} &= \mathcal{X}_{i}^{l} \odot \mathcal{M} + \mathcal{X}_{j}^{l} \odot (1 - \mathcal{M}) \\
  \mathcal{Y}_{syn(i,j)}^{l} &= \mathcal{Y}_{i}^{l} \odot \mathcal{M} + \mathcal{Y}_{j}^{l} \odot (1 - \mathcal{M}),
\end{aligned}
\end{equation}
where $\mathcal{X}_{syn(i,j)}^{l}$ and $\mathcal{Y}_{syn(i,j)}^{l}$ are synthetic images and labels for Task\#5. $\mathcal{M}$ is a mask with 30\% $\sim$ 70\% random masked regions. Symbol $\odot$ is element-wise multiplication. Note that $\mathcal{Y}_{i}^{l}$ and $\mathcal{Y}_{j}^{l}$ are binary masks, $\mathcal{X}_{i}^{l}$ and $\mathcal{X}_{j}^{l}$ are images from the $i$-th and $j$-th task, thus $\mathcal{X}_{syn(i,j)}^{l}$ can be regarded as mixed data that contain various targets and background. For labeled data learning (containing Task\#5), Dice loss and cross-entropy loss are leveraged, defined as:
\begin{equation}
\begin{split}
  \mathcal{L}_{lab} = Dice(\mathcal{F}(\mathcal{X}_{k}^{l}, [Prompt_{\#k}]; \Theta), \mathcal{Y}_{k}^{l}) + \\
  CE(\mathcal{F}(\mathcal{X}_{k}^{l}, [Prompt_{\#k}]; \Theta), \mathcal{Y}_{k}^{l}),
  \label{eq:}
\end{split}
\end{equation}
where $\mathcal{L}_{lab}$ is the supervised loss on labeled data. For simplicity, we use $\mathcal{F}(\cdot;\Theta)$ to define the whole network with parameter $\Theta$, which contains operations in Eq.\ref{eq:one}. Benefited by Task\#5, VerSemi could have a semantic perception of all other segmentation tasks.


{\noindent\bf Enhancing the controllability of task prompt by $\mathcal{L}_{aux}$.} As indicated by Fig.~\ref{dod_uni}, there exists a weakening phenomenon of task prompt when using task-prompted dynamic head, in which we can find models~(\eg, DoDNet~\cite{zhang2021dodnet}) sometimes fail to recognize prescriptive task even under the control of task-specific prompt. It might be caused by the shared semantic information between different segmentation tasks. Therefore, to enhance the uniqueness of the task prompt, we add an auxiliary constraint $\mathcal{L}_{aux}$ formulated as:
\begin{equation}
\begin{split}
  \mathcal{L}_{aux} = Dice(\mathcal{F}(\mathcal{X}_{syn(i,j)}^{l}, [Prompt_{\#k}]; \Theta), \mathcal{Y}_{k}^{l}) + \\
  CE(\mathcal{F}(\mathcal{X}_{syn(i,j)}^{l}, [Prompt_{\#k}]; \Theta), \mathcal{Y}_{k}^{l}) k=i,j.
  \label{eq:}
\end{split}
\end{equation}

This formula can be concluded as follows: models can only show interest in the specifically prompted task, when facing mixed data. So far, the supervised loss $\mathcal{L}_{sup}$ is written as:
\begin{equation}
  \mathcal{L}_{sup} = \mathcal{L}_{lab} + \mathcal{L}_{aux}.
  \label{eq:}
\end{equation}

In this way, VerSemi devises a synthetic Task\#5, which paves the way for task-agnostic learning from the unlabeled data, as well as augmenting the capability of task prompt with the help of $\mathcal{L}_{aux}$.

\begin{table*}[t]
\caption{Performance comparison on the pancreas dataset, in the scenario of leveraging 10\% and 20\% labeled data. The best and second best results are shown in \textcolor{red}{red} and \textcolor{blue}{blue}, respectively. ((Dice, \%); (Jaccard, \%); (ASD, voxel); (95HD, voxel).)}\label{panc_res}

\centering 
\resizebox{0.85\linewidth}{!}
{
\begin{tabular}{l|cccc|cccc}
\hline
\multirow{2}{*}{Method} 
& \multicolumn{4}{c|}{Pancreas (10\%/6 labeled data)}          
& \multicolumn{4}{c}{Pancreas (20\%/12 labeled data)}      \\\cline{2-9} 
& \multicolumn{1}{c}{Dice~$\uparrow$} & \multicolumn{1}{c}{Jaccard~$\uparrow$} & \multicolumn{1}{c}{ASD~$\downarrow$} & 95HD~$\downarrow$ & \multicolumn{1}{c}{Dice~$\uparrow$} & \multicolumn{1}{c}{Jaccard~$\uparrow$} & \multicolumn{1}{c}{ASD~$\downarrow$} & 95HD~$\downarrow$ \\ \hline
VNet~(3DV'16)~\cite{milletari2016v}  & \multicolumn{1}{c}{55.60}     & \multicolumn{1}{c}{41.74}        & \multicolumn{1}{c}{18.63}    & 45.33     & \multicolumn{1}{c}{72.38}     & \multicolumn{1}{c}{58.26}    & \multicolumn{1}{c}{5.89}    & 19.35 \\ \hline
UA-MT~(MICCAI'19)~\cite{yu2019uncertainty}  & \multicolumn{1}{c}{66.34}     & \multicolumn{1}{c}{53.21}        & \multicolumn{1}{c}{4.57}    & 17.21     & \multicolumn{1}{c}{76.10}     & \multicolumn{1}{c}{62.62}     & \multicolumn{1}{c}{2.43}   & 10.84     \\ 
DTC~(AAAI'19)~\cite{luo2021semi} & \multicolumn{1}{c}{69.21}     & \multicolumn{1}{c}{54.06}        & \multicolumn{1}{c}{5.95}    &  17.21    & \multicolumn{1}{c}{78.27}     & \multicolumn{1}{c}{64.75}  & \multicolumn{1}{c}{2.25}    &  8.36   \\ 
ASE-Net~(TMI'22)~\cite{lei2022semi} & \multicolumn{1}{c}{71.54}     & \multicolumn{1}{c}{56.82}        & \multicolumn{1}{c}{5.73}    &  16.33    & \multicolumn{1}{c}{79.03}     & \multicolumn{1}{c}{66.57}  & \multicolumn{1}{c}{2.30}    &  8.62   \\ 
CAML~(MICCAI'23)~\cite{gao2023correlation} & \multicolumn{1}{c}{71.21}     & \multicolumn{1}{c}{56.32}        & \multicolumn{1}{c}{5.92}    &  {16.89}    & \multicolumn{1}{c}{79.81}     & \multicolumn{1}{c}{67.35}  & \multicolumn{1}{c}{2.27}    &  {8.22}   \\ 
BCP~(CVPR'23)~\cite{bai2023bidirectional} & \multicolumn{1}{c}{73.83}     & \multicolumn{1}{c}{59.24}        & \multicolumn{1}{c}{3.72}    &  \textcolor{blue}{12.71}    & \multicolumn{1}{c}{\textcolor{blue}{82.91}}     & \multicolumn{1}{c}{\textcolor{blue}{70.97}}  & \multicolumn{1}{c}{\textcolor{blue}{2.25}}    &  \textcolor{blue}{6.43}   \\ 
CauSSL~(ICCV'23)~\cite{miao2023caussl}  & \multicolumn{1}{c}{72.34}     & \multicolumn{1}{c}{57.43}        & \multicolumn{1}{c}{\textcolor{blue}{3.13}}    &  {13.49}    & \multicolumn{1}{c}{80.63}     & \multicolumn{1}{c}{67.84}  & \multicolumn{1}{c}{2.78}    &  {8.76}   \\ 
MagicNet~(CVPR'23)~\cite{chen2023magicnet}  & \multicolumn{1}{c}{\textcolor{blue}{75.01}}     & \multicolumn{1}{c}{\textcolor{blue}{62.04}}        & \multicolumn{1}{c}{3.97}    &  {13.71}    & \multicolumn{1}{c}{81.25}     & \multicolumn{1}{c}{68.81}  & \multicolumn{1}{c}{2.83}    &  {8.50}   \\ \hline
VerSemi  & \multicolumn{1}{c}{\textcolor{red}{78.08}}     & \multicolumn{1}{c}{\textcolor{red}{64.82}}        & \multicolumn{1}{c}{\textcolor{red}{2.33}}    &  \textcolor{red}{{8.05}}    & \multicolumn{1}{c}{\textcolor{red}{83.27}}     & \multicolumn{1}{c}{\textcolor{red}{71.68}}  & \multicolumn{1}{c}{\textcolor{red}{1.40}}    &  {\textcolor{red}{5.33}}   \\ \hline
\rowcolor{gray!15} VerSemi w/ Task Info & \multicolumn{1}{c}{78.62}     & \multicolumn{1}{c}{64.91}        & \multicolumn{1}{c}{2.28}    &  {7.99}    & \multicolumn{1}{c}{83.55}     & \multicolumn{1}{c}{71.93}  & \multicolumn{1}{c}{1.35}    &  {5.02}   \\ \hline
\end{tabular}
}
\end{table*}

\begin{table*}[t]
\caption{Performance comparison on the spleen dataset, in the scenario of leveraging 10\% and 20\% labeled data. The best and second best results are shown in \textcolor{red}{red} and \textcolor{blue}{blue}, respectively. ((Dice, \%); (Jaccard, \%); (ASD, voxel); (95HD, voxel).)}\label{sp_res}

\centering 
\resizebox{0.85\linewidth}{!}
{
\begin{tabular}{l|cccc|cccc}
\hline
\multirow{2}{*}{Method} 
& \multicolumn{4}{c|}{Spleen (10\%/3 labeled data)}          
& \multicolumn{4}{c}{Spleen (20\%/6 labeled data)}      \\\cline{2-9} 
& \multicolumn{1}{c}{Dice~$\uparrow$} & \multicolumn{1}{c}{Jaccard~$\uparrow$} & \multicolumn{1}{c}{ASD~$\downarrow$} & 95HD~$\downarrow$ & \multicolumn{1}{c}{Dice~$\uparrow$} & \multicolumn{1}{c}{Jaccard~$\uparrow$} & \multicolumn{1}{c}{ASD~$\downarrow$} & 95HD~$\downarrow$ \\ \hline
VNet~(3DV'16)~\cite{milletari2016v} & \multicolumn{1}{c}{75.14}     & \multicolumn{1}{c}{65.27}        & \multicolumn{1}{c}{15.02}    & 43.89     & \multicolumn{1}{c}{79.78}     & \multicolumn{1}{c}{72.86}    & \multicolumn{1}{c}{11.37}    & 30.03 \\ \hline
UA-MT~(MICCAI'19)~\cite{yu2019uncertainty}  & \multicolumn{1}{c}{79.63}     & \multicolumn{1}{c}{68.62}        & \multicolumn{1}{c}{15.94}    & {44.71}    & \multicolumn{1}{c}{83.11}     & \multicolumn{1}{c}{75.98}     & \multicolumn{1}{c}{8.92}   & {25.41}     \\ 
DTC~(AAAI'19)~\cite{luo2021semi}  & \multicolumn{1}{c}{80.27}     & \multicolumn{1}{c}{69.00}        & \multicolumn{1}{c}{14.53}    &  {41.56}    & \multicolumn{1}{c}{84.59}     & \multicolumn{1}{c}{75.91}  & \multicolumn{1}{c}{9.75}    &  {31.77}    \\ 
ASE-Net~(TMI'22)~\cite{lei2022semi}  & \multicolumn{1}{c}{80.65}     & \multicolumn{1}{c}{69.48}        & \multicolumn{1}{c}{14.37}    &  \textcolor{blue}{41.31}    & \multicolumn{1}{c}{85.02}     & \multicolumn{1}{c}{75.68}  & \multicolumn{1}{c}{12.53}    &  {37.26}   \\ 
CAML~(MICCAI'23)~\cite{gao2023correlation}  & \multicolumn{1}{c}{80.32}     & \multicolumn{1}{c}{69.10}        & \multicolumn{1}{c}{15.37}    &  41.71    & \multicolumn{1}{c}{{85.80}}     & \multicolumn{1}{c}{76.79}  & \multicolumn{1}{c}{11.57}    & 36.14   \\ 
BCP~(CVPR'23)~\cite{bai2023bidirectional}  & \multicolumn{1}{c}{83.12}     & \multicolumn{1}{c}{72.85}        & \multicolumn{1}{c}{14.42}    &  {42.11}    & \multicolumn{1}{c}{{87.02}}     & \multicolumn{1}{c}{78.58}  & \multicolumn{1}{c}{10.48}    & 37.08   \\ 
CauSSL~(ICCV'23)~\cite{miao2023caussl} & \multicolumn{1}{c}{81.98}     & \multicolumn{1}{c}{71.25}        & \multicolumn{1}{c}{14.69}    &  {41.84}    & \multicolumn{1}{c}{86.83}     & \multicolumn{1}{c}{78.46}  & \multicolumn{1}{c}{10.01}    &  {32.27}   \\ 
MagicNet~(CVPR'23)~\cite{chen2023magicnet}  & \multicolumn{1}{c}{\textcolor{blue}{83.55}}     & \multicolumn{1}{c}{\textcolor{blue}{73.58}}        & \multicolumn{1}{c}{\textcolor{blue}{13.49}}    &  {41.79}    & \multicolumn{1}{c}{\textcolor{blue}{88.24}}     & \multicolumn{1}{c}{\textcolor{blue}{80.24}}  & \multicolumn{1}{c}{\textcolor{blue}{8.50}}    &  \textcolor{blue}{23.51}  \\ \hline
VerSemi  & \multicolumn{1}{c}{\textcolor{red}{89.34}}     & \multicolumn{1}{c}{\textcolor{red}{81.73}}        & \multicolumn{1}{c}{\textcolor{red}{3.12}}    &  \textcolor{red}{9.33}    & \multicolumn{1}{c}{\textcolor{red}{94.62}}     & \multicolumn{1}{c}{\textcolor{red}{89.89}}  & \multicolumn{1}{c}{\textcolor{red}{2.40}}    &  \textcolor{red}{7.50}   \\ \hline
\rowcolor{gray!15} VerSemi w/ Task Info  & \multicolumn{1}{c}{90.10}     & \multicolumn{1}{c}{82.75}        & \multicolumn{1}{c}{3.09}    &  {9.28}    & \multicolumn{1}{c}{94.67}     & \multicolumn{1}{c}{89.93}  & \multicolumn{1}{c}{2.35}    &  {7.33}  \\ \hline
\end{tabular}
}
\end{table*}

\subsection{Task-agnostic Unlabeled Data Learning}
\label{sec3.3}

Considering task specifics are required for prompt-driven model to generate prompt, here we place our VerSemi in a more demanding SSL context, in which unlabeled task specifics are not desired. Below we describe how this is achieved. Firstly, CutMix~\cite{yun2019cutmix} is conducted on all unlabeled data, making the input contain objects of different tasks. Then the prediction with \textbf{Task\#5 Prompt} is forced to be consistent with the aggregated prediction using \textbf{Task\#1 Prompt} $\sim$ \textbf{Task\#4 Prompt}~(see Fig.~\ref{unlabeled_learning}). The aggregated prediction can be regarded as a combination of pseudo-masks for each task, and the prediction prompted by Task\#5 can be considered as a direct pseudo-mask for all tasks. Therefore, the two predictions should be identical. We call this operation self-consistency since no extra decoder or teacher model is required for supervision. The entire process can be written as:
\begin{equation}
\begin{aligned}
  \mathcal{X}_{syn(i,j)}^{u} = \mathcal{X}_{i}^{u} &\odot \mathcal{M} + \mathcal{X}_{j}^{u} \odot (1 - \mathcal{M}) \\
  \mathcal{P}_{agg} = \max\limits_{k\in(1,4)} (\mathcal{F}(&\mathcal{X}_{syn(i,j)}^{u}, [Prompt_{\#k}]; \Theta)),
\end{aligned}
\end{equation}
where $\mathcal{X}_{syn(i,j)}^{u}$ are mixed unlabeled data, $\mathcal{X}_{i}^{u}$ and $\mathcal{X}_{j}^{u}$ are randomly selected unlabeled data. 
Element-wise maximization is performed to aggregate predictions prompted by Task\#1 $\sim$ Task\#4, and $\mathcal{P}_{agg}$ is the final aggregated prediction. The overall loss $\mathcal{L}_{total}$ and unsupervised loss $\mathcal{L}_{unsup}$ are calculated as:
\begin{equation}
\begin{aligned}
   \mathcal{L}_{total} = &\mathcal{L}_{sup} + \mathcal{L}_{unsup} \\
 \mathcal{L}_{unsup} = Dice(\mathcal{P}_{agg}, &\mathcal{F}(\mathcal{X}_{syn(i,j)}^{u}, [Prompt_{\#5}]; \Theta)).
\end{aligned}
\end{equation}

To summarize, based on the design of semantic-aware Task\#5, our VerSemi learns from unlabeled data in a task-agnostic way, and also enhances the the uniqueness of the task prompt with the auxiliary constraint $\mathcal{L}_{aux}$.

\section{Experiments and Results}

\subsection{Setup}

{\noindent\bf Datasets. }We report the model segmentation results on four public datasets, including \textbf{Task\#1:} NIH-Pancreas~\cite{roth2015deeporgan}, \textbf{Task\#2:} Left Atrium~\cite{xiong2021global}, \textbf{Task\#3:} MSD-Spleen~\cite{antonelli2022medical} and \textbf{Task\#4:} MSD-Lung Tumor~\cite{antonelli2022medical}. Specifically, NIH-Pancreas contains 82 contrast-enhanced abdomen CT scans, which are split into 62/20 scans for training/test. The Left Atrium has 100 gadolinium-enhanced MR images, in which 80/20 images are leveraged for training/test. MSD-Spleen contains 41 CT scans, and 30/11 scans are split for training/test. MSD-Lung Tumor contains 63 CT scans, which are divided into 50/13 scans for training/test. Among them, 10\% training data are split into a validation set to select the best model. All methods follow the same data split for fair comparisons, with the same pre-processing as~\cite{luo2021semi, bai2023bidirectional}.

{\noindent\bf Implementation Details. }Following previous works~\cite{bai2023bidirectional, luo2021semi, chen2023magicnet}, we adopt V-Net~\cite{milletari2016v} as the baseline model for fair comparisons. We use the Adam optimizer~\cite{kingma2014adam} with a learning rate of 0.001. The input size and batch size are set to 96$\times$96$\times$96 and 8, respectively. Experiments were implemented by Pytorch~\cite{paszke2019pytorch} with four NVIDIA GeForce RTX 3080 Ti GPUs. Evaluation metrics of Dice~(\%), Jaccard~(\%), Average Surface Distance (ASD, voxel) and 95\% Hausdorff Distance (95HD, voxel) are used here.

\subsection{Comparison with Existing Methods}
We compare our VerSemi with seven popular SSL methods, including uncertainty-aware mean-teacher (UA-MT)~\cite{yu2019uncertainty}, dual-task consistency (DTC)~\cite{luo2021semi}, adversarial consistency and dynamic convolution (ASE-Net)~\cite{lei2022semi}, correlation-aware mutual learning (CAML)~\cite{gao2023correlation}, bidirectional copy-paste~\cite{bai2023bidirectional}, causality-inspired semi-supervised segmentation (CauSSL)~\cite{miao2023caussl} and cubic volume partition and recovery (Magic-Net)~\cite{chen2023magicnet}. (Results on Left Atrium dataset and Lung Tumor dataset are provided in the supplementary)

\begin{figure}[t]
    \centering
    \includegraphics[width=1.0\linewidth]{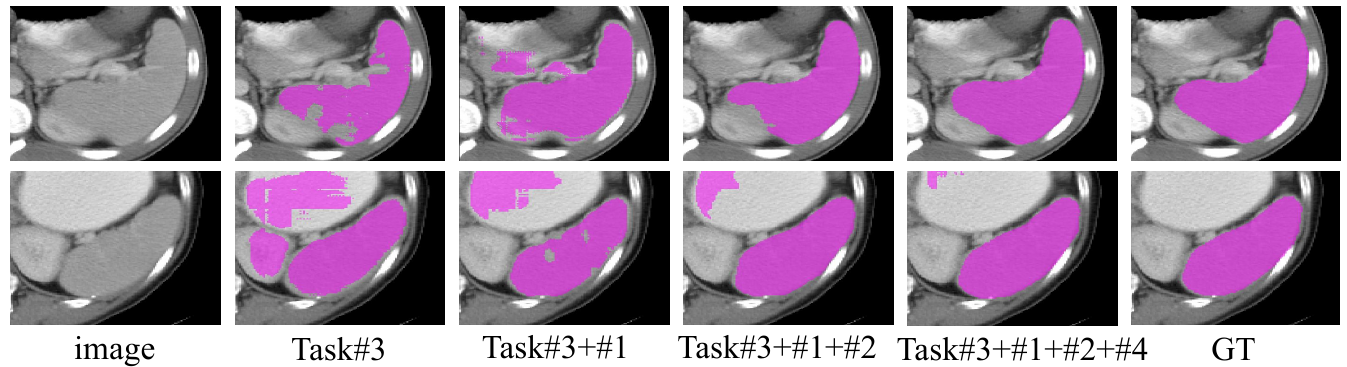}
    \caption{Visualization results of spleen segmentation by incorporating other tasks sequentially. We can find the model tends to produce more accurate segmentation with the increase of integrated tasks, demonstrating the benefits of learning a unified model.}
    \label{abla_vis}
\end{figure}

\begin{table}[t]  
  \setlength{\tabcolsep}{1.60mm}
  \centering  
  \caption{A case study of the impact of other tasks on one specific task. Here spleen segmentation task~(Task \#3) is selected as the baseline, as we find VerSemi presents remarkable improvements on this task when compared to other methods, and this experiment aims to figure out where the performance gains come from.} \label{abla}
  \begin{threeparttable}  
    \begin{tabular}{lccccccc}
    \toprule  
    \multirow{2}{*}{Setting} 
    &\multicolumn{2}{c}{10\% labels} &\multicolumn{2}{c}{20\% labels}  \cr 
    \cmidrule(lr){2-3} \cmidrule(lr){4-5}   
    {} & Dice~$\uparrow$ & 95HD~$\downarrow$ & Dice~$\uparrow$ & 95HD~$\downarrow$ \cr  
    \midrule
    {Task\#3} & {75.14} & {43.89} & {79.78} & {30.03} \cr
    {Task\#3+\#1} & {85.62} & {17.07} & {90.00} & {15.81} \cr
    {Task\#3+\#1+\#2} & {88.03} & {11.06} & {92.06} & {10.06}  \cr
    {Task\#3+\#1+\#2+\#4} & \textbf{89.34} & \textbf{9.33} & \textbf{94.62} & \textbf{7.50} \cr
    \bottomrule 
    \end{tabular} 
    \end{threeparttable}  
    \label{table5}
\end{table}

{\noindent\bf Results on Pancreas Dataset. }As shown in Table~\ref{panc_res}, we can find VerSemi consistently surpasses others on all metrics under different SSL settings. For example, VerSemi brings respectively 3.07\% and 5.66~(voxels) improvements on Dice and HD score than the second best method MagicNet with 10\% labeled data training. Besides, we can find the performance gains obtained by VerSemi are larger than the others when leveraging fewer labels, indicating the effectiveness of our VerSemi in annotation-scarce scenarios.

\begin{table*}[t]
\caption{Adapting BCP~\cite{bai2023bidirectional} and CauSSL~\cite{miao2023caussl} into unified SSL models. However, severe performance degradation can be seen when comparing Uni-BCP to BCP and Uni-CauSSL to CauSSL.}\label{other_ssl}

\centering 
\resizebox{0.85\linewidth}{!}
{
\begin{tabular}{c|cccc|cccc}
\hline
\multirow{2}{*}{Method} 
& \multicolumn{4}{c|}{Pancreas (10\%/6 labeled data)}          
& \multicolumn{4}{c}{Left Atrium (10\%/8 labeled data)}      \\\cline{2-9} 
& \multicolumn{1}{c}{Dice~$\uparrow$} & \multicolumn{1}{c}{Jaccard~$\uparrow$} & \multicolumn{1}{c}{ASD~$\downarrow$} & 95HD~$\downarrow$ & \multicolumn{1}{c}{Dice~$\uparrow$} & \multicolumn{1}{c}{Jaccard~$\uparrow$} & \multicolumn{1}{c}{ASD~$\downarrow$} & 95HD~$\downarrow$ \\ \hline
Uni-BCP& \multicolumn{1}{c}{68.59}     & \multicolumn{1}{c}{53.73}        & \multicolumn{1}{c}{7.33}    &  {20.62}    & \multicolumn{1}{c}{85.73}     & \multicolumn{1}{c}{75.06}  & \multicolumn{1}{c}{10.17}    & {30.33}   \\ 
Uni-CauSSL& \multicolumn{1}{c}{65.35}     & \multicolumn{1}{c}{49.09}        & \multicolumn{1}{c}{6.16}    &  {20.89}    & \multicolumn{1}{c}{83.40}     & \multicolumn{1}{c}{72.43}  & \multicolumn{1}{c}{8.84}    &  {34.94}   \\ 
VerSemi w/o $\mathcal{L}_{aux}$ & \multicolumn{1}{c}{75.06}     & \multicolumn{1}{c}{60.94}        & \multicolumn{1}{c}{3.70}    &  {11.64}    & \multicolumn{1}{c}{88.56}     & \multicolumn{1}{c}{79.81}  & \multicolumn{1}{c}{2.62}    &  {9.17}   \\ 
VerSemi~(Ours)& \multicolumn{1}{c}{\textbf{78.08}}     & \multicolumn{1}{c}{\textbf{64.82}}        & \multicolumn{1}{c}{\textbf{2.33}}    &  \textbf{8.05}    & \multicolumn{1}{c}{\textbf{89.01}}     & \multicolumn{1}{c}{\textbf{80.52}}  & \multicolumn{1}{c}{\textbf{2.57}}    & \textbf{9.03}   \\ \hline\hline
\multirow{2}{*}{Method} 
& \multicolumn{4}{c|}{Spleen (10\%/3 labeled data)}          
& \multicolumn{4}{c}{Lung Tumor (10\%/5 labeled data)}      \\\cline{2-9} 
& \multicolumn{1}{c}{Dice~$\uparrow$} & \multicolumn{1}{c}{Jaccard~$\uparrow$} & \multicolumn{1}{c}{ASD~$\downarrow$} & 95HD~$\downarrow$ & \multicolumn{1}{c}{Dice~$\uparrow$} & \multicolumn{1}{c}{Jaccard~$\uparrow$} & \multicolumn{1}{c}{ASD~$\downarrow$} & 95HD~$\downarrow$ \\ \hline
Uni-BCP& \multicolumn{1}{c}{74.80}     & \multicolumn{1}{c}{58.89}        & \multicolumn{1}{c}{17.11}    &  {54.06}    & \multicolumn{1}{c}{31.01}     & \multicolumn{1}{c}{21.32}  & \multicolumn{1}{c}{11.35}    & {24.36}   \\ 
Uni-CauSSL& \multicolumn{1}{c}{73.06}     & \multicolumn{1}{c}{57.85}        & \multicolumn{1}{c}{18.28}    &  {55.51}    & \multicolumn{1}{c}{25.38}     & \multicolumn{1}{c}{20.20}  & \multicolumn{1}{c}{15.06}    &  {28.72}   \\ 
VerSemi w/o $\mathcal{L}_{aux}$ & \multicolumn{1}{c}{87.80}     & \multicolumn{1}{c}{79.19}        & \multicolumn{1}{c}{3.36}    &  {10.10}    & \multicolumn{1}{c}{34.74}     & \multicolumn{1}{c}{22.55}  & \multicolumn{1}{c}{12.62}    &  {24.77}   \\ 
VerSemi~(Ours)& \multicolumn{1}{c}{\textbf{89.34}}     & \multicolumn{1}{c}{\textbf{81.73}}        & \multicolumn{1}{c}{\textbf{3.12}}    &  \textbf{9.33}    & \multicolumn{1}{c}{\textbf{36.90}}     & \multicolumn{1}{c}{\textbf{28.12}}  & \multicolumn{1}{c}{\textbf{10.87}}    & \textbf{23.41}   \\ \hline
\end{tabular}
}
\end{table*}

\begin{figure}[t]
    \centering
    \includegraphics[width=1.0\linewidth]{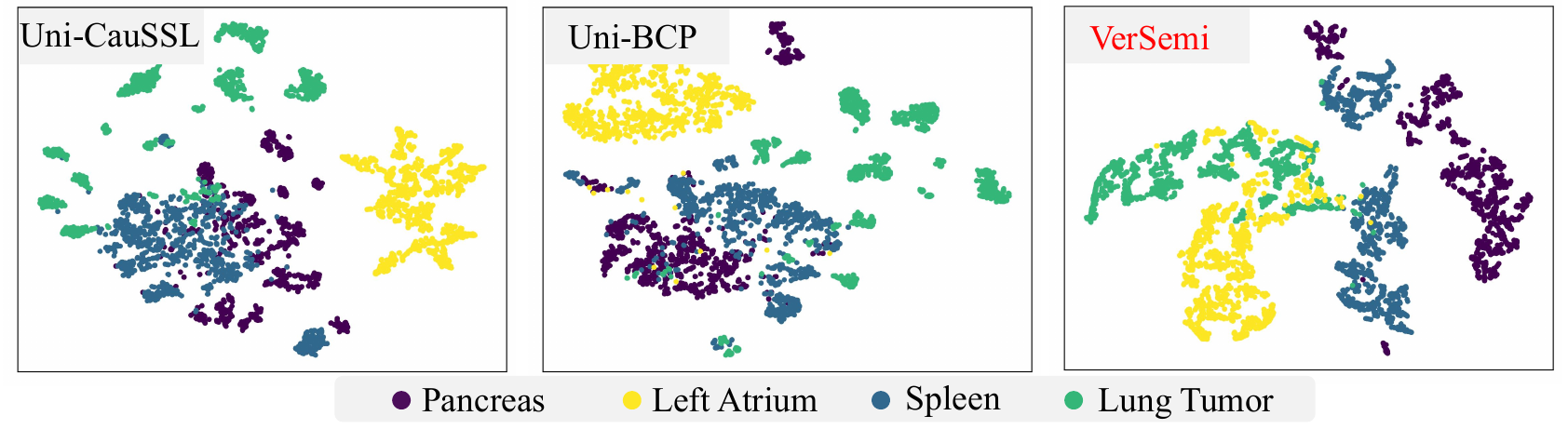}
    \caption{t-SNE visualization of feature embedding for four tasks. The implemented Uni-CauSSL, Uni-BCP and our proposed VerSemi are compared.}
    \label{tsne}
\end{figure}

{\noindent\bf Results on Spleen Dataset. }Table~\ref{sp_res} presents the results of spleen segmentation. We can see that our VerSemi significantly outperforms all other competitors by a large margin. For instance, compared to MagicNet, VerSemi has respectively 32.46~(voxels) and 16.01~(voxels) performance gains on HD score with 10\% and 20\% labeled data. Similarly, VerSemi surpasses CauSSL by 7.36\% on Dice score under 10\% label percentage. 
A case study is conducted to see the performance gains by introducing other tasks on spleen segmentation (Task\#3). As Table~\ref{abla} and Fig.~\ref{abla_vis} show, there are consistent improvements by gradually integrating data of other tasks. In particular, we can find the performance improved greatly by integrating the pancreas segmentation (Task\#1), where our model has already outperformed other methods. Here we provide three potential reasons for VerSemi's high performance on this task: (1) due to extremely limited labels (10\% labels are equal to 3 labeled data), competitors fail to generalize the representation learned from labeled data to unlabeled data, and mistakenly predict the background as foreground (see Row 5-6 of Fig.\ref{seg_vis}). Therefore, a high HD score can be observed; (2) since the same modality, $e.g.$, Task\#1 and Task\#3, VerSemi can learn modality-specific knowledge and achieve better performance. (\eg, see Fig.~\ref{tsne}, the feature embedding of pancreas and spleen are very close in the latent space, both of them are abdominal organs); and (3) by segmenting other organs, VerSemi can segment the background regions and identify the adhesive boundaries in a negative learning mechanism, so as to decrease the HD score.

\subsection{In-depth Analysis}

{\noindent\bf Importance of the auxiliary constraint $\mathcal{L}_{aux}$. }$\mathcal{L}_{aux}$ plays the role of augmenting the uniqueness of task prompts. As the last two rows of Table~\ref{other_ssl} indicates, by incorporating $\mathcal{L}_{aux}$, VerSemi presents respectively 3.02\%, 0.45\%, 1.54\% and 2.16\% performance gains on Dice score on the pancreas, left atrium, spleen and lung tumor tasks, when using 10\% labeled data. This improvement demonstrates the effectiveness and necessity of adding an accessory loss to constrain the feasibility of task prompts.

\begin{figure}[t]
    \centering
    \includegraphics[width=0.4\textwidth]{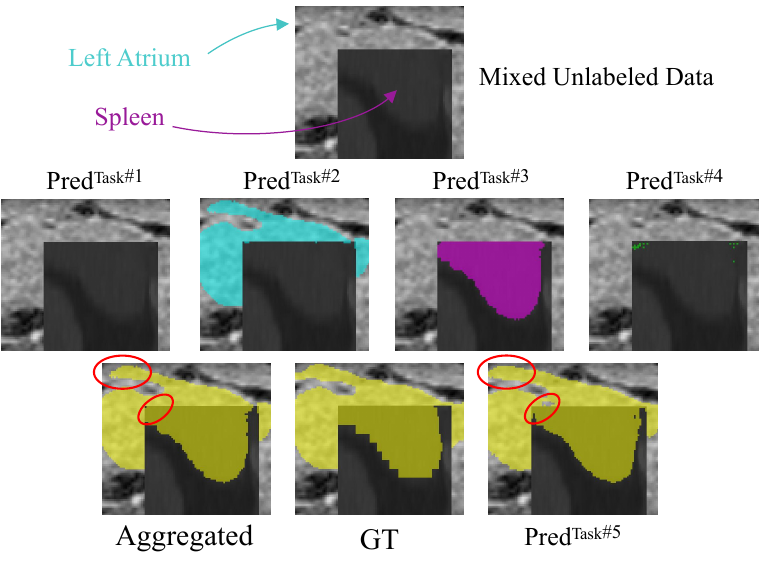}
    \caption{Visualization results prompted by pertinent tasks and Task\#5, when meeting mixed unlabeled data. In this case, the spleen and left atrium are mixed together. The inconsistent regions between aggregated prediction and Task\#5-prompted prediction are highlighted by \textcolor{red}{red elliptic}.}
    \label{unlabeled_learning}
    \vspace{-5mm}
\end{figure}

{\noindent\bf Adapting single SSL models into unified SSL models. }In this experiment, we revise CauSSL and BCP into the unified SSL settings. There are two changes compared to their previous versions. (1) The input data cover four tasks and are randomly fed into the model with the associated task id. (2) Changing the number of output channels to match the number of tasks, which is different from VerSemi as VerSemi has a dynamic task-prompted head with two output channels. As Table~\ref{other_ssl} shows, the results produced by Uni-BCP and Uni-CauSSL are far inferior to VerSemi, and compared to their original single model version, significant performance degradation can be observed. For instance, according to the averaged Dice score with 10\% labeled data, BCP vs Uni-BCP (70.79\% vs 65.03\%), 5.76\% drop can be found. And CauSSL vs Uni-CauSSL (69.60\% vs 61.80\%), 7.80\% degradation is discovered. This phenomenon is mainly triggered by chaotic representation learned from all task data, and also indicates that naively learning from all tasks simultaneously is not effective and even harmful to the single task. Moreover, we plot the t-SNE visualization of feature embedding to have a clear view. As Fig.~\ref{tsne} exhibits, VerSemi presents a distinguishable decision boundary while others show mixed and dispersed embedding. 
This demonstrates that task prompts and the constraint to task prompts~($\mathcal{L}_{aux}$) are essential when facing multiple SSL tasks, as the former guides model to have a clear understanding of the ongoing task, while the latter makes sure the learned representation of each single task are discernible and concentrated.

\begin{figure}[t]
    \centering
    \includegraphics[width=1.0\linewidth]{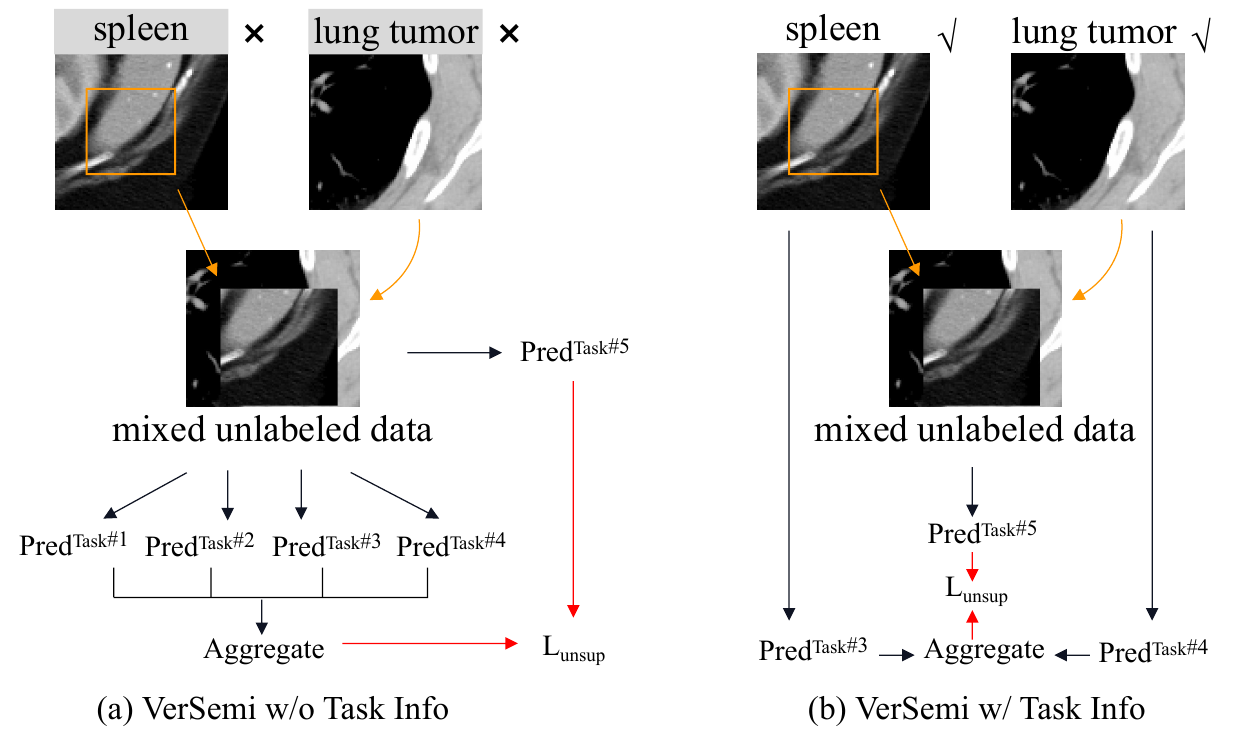}
    \caption{(a) VerSemi is designed to learn from unlabeled data without knowing associated task info. (b) The pipeline of VerSemi when unlabeled task info is given.}
    \label{task_info}
\end{figure}

{\noindent\bf Visualization of unlabeled data learning pipeline. }Fig.~\ref{unlabeled_learning} shows the segmentation results prompted by pertinent tasks and synthetic Task\#5. We can see that VerSemi can clearly recognize the task-specific prompted regions, which is largely benefited by the auxiliary constraint $\mathcal{L}_{aux}$, for its ability to enhance the controllability of prompts. Meanwhile, we can also find VerSemi smoothly highlights all task semantic regions under the prompt of Task\#5, demonstrating the effectiveness of learning a semantic-aware synthetic task. By aligning the two predictions~(see the bottom of Fig.~\ref{unlabeled_learning}, Aggregated and Perd$^{Task\#5}$), VerSemi learns the unlabeled data in a task-agnostic manner.

{\noindent\bf Incorporating unlabeled task information into VerSemi. }To explore the upper bound of VerSemi, we report the results when feeding task information of unlabeled data into VerSemi. As Fig.~\ref{task_info} presents, VerSemi w/ task info can directly generate predictions on the source image with task-specific prompts, whereas VerSemi w/o task info should first generate predictions on the mixed data with all pertinent task prompts and then aggregate them. From the last-row results of Table~\ref{panc_res} and Table~\ref{sp_res} with \colorbox {gray!15}{gray} background, we can find there is an improvement (\ie, a 0.78\% gain on the averaged Dice score with 10\% labels), demonstrating the ability to produce accurate predictions with mixed data, as well as distinguishing task-prompted specific regions.

\begin{table}[t]  
  \setlength{\tabcolsep}{1.52mm}
  \centering  
  \caption{Discussion of three types of prompt, in which language prompt, soft and one-hot vector prompts are compared. The averaged Dice and 95HD scores \textbf{on four tasks} are reported.} \label{task_prompt}
  \begin{threeparttable}  
    \begin{tabular}{lccccccc}
    \toprule  
    \multirow{2}{*}{types of prompt} 
    &\multicolumn{2}{c}{10\% labels} &\multicolumn{2}{c}{20\% labels}  \cr 
    \cmidrule(lr){2-3} \cmidrule(lr){4-5}   
    {} & Dice~$\uparrow$ & 95HD~$\downarrow$ & Dice~$\uparrow$ & 95HD~$\downarrow$ \cr  
    \midrule
    {language~(CLIP)} & {67.21}  &  {22.78} & {74.30} & {19.29} \cr
    {soft vector} & {70.02} & {17.66} & {77.45} & {13.83} \cr
    {one-hot vector} & \textbf{73.33} & \textbf{12.46} & \textbf{80.99} & \textbf{8.74} \cr
    \bottomrule 
    \end{tabular} 
    \end{threeparttable}
\end{table}


{\noindent\bf Discussion of task prompt. }Prompt is used as a signal to help the model understand the ongoing task, typically language~\cite{yao2023visual, liu2023clip}~(using a sentence to describe), soft vector~\cite{vu2021spot, ye2023uniseg}~(using randomly initialized learnable vector to represent) and one-hot prompts~\cite{zhang2021dodnet, yasunaga2022linkbert} are mostly employed. As Table~\ref{task_prompt} shows, the one-hot prompt performs best under SSL setting. Reasons for the results are: (1) the embedding of language heavily relies on language or vision-language models, which is not guaranteed to be aligned with the extracted medical image embedding; (2) soft vector prompt works when there are substantial paired image-label data, whereas only scarce labels are available in the context of SSL, making it hard to adapt. By contrast, the one-hot prompt is more explicit and empirically suitable for SSL.



\begin{figure}[!t]
    \centering
    \includegraphics[width=1.0\linewidth]{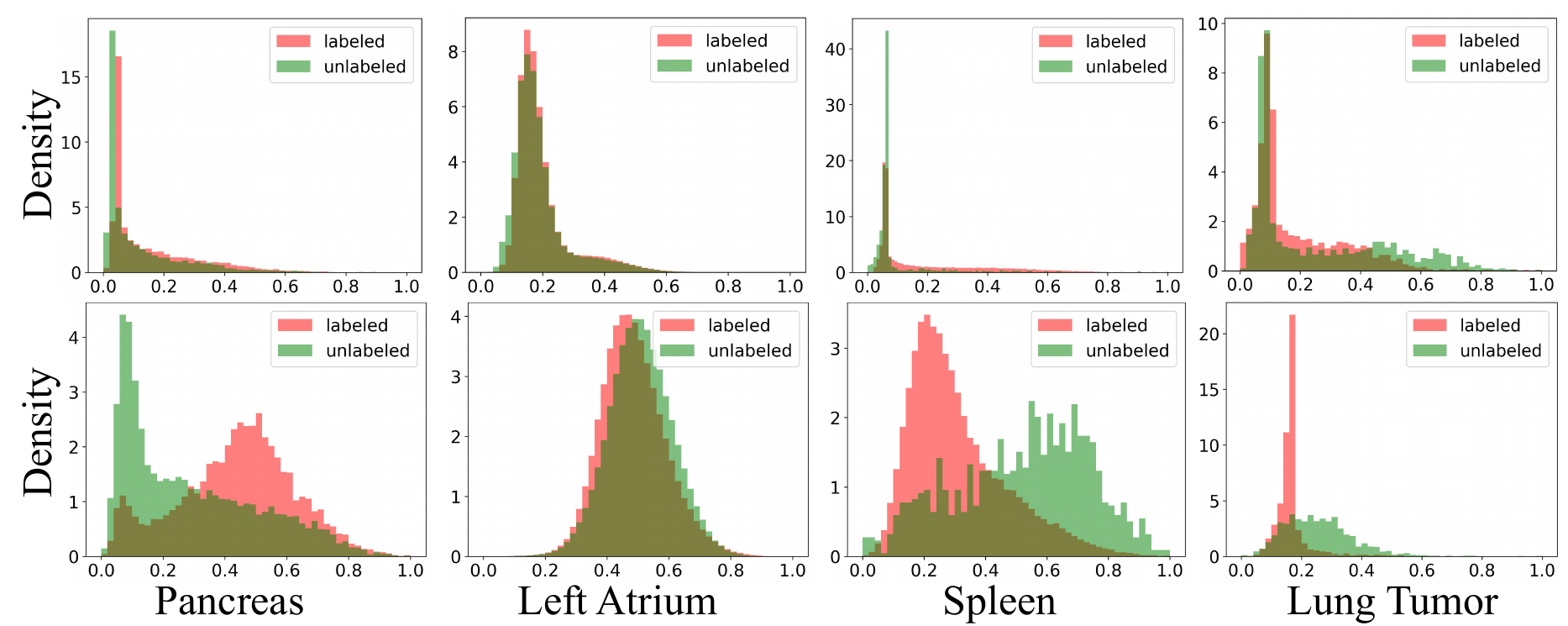}
    \caption{Kernel density estimation of VerSemi and BCP when training with 10\% labels. This experiment is conducted to see the distribution of labeled and unlabeled data. \textbf{Top} is the distribution generated by VerSemi and \textbf{Bottom} is BCP~\citep{bai2023bidirectional}. It is clear to see that VerSemi aligns the distribution better than BCP.}
    \label{kde}
    \vspace{-3mm}
\end{figure}

\begin{figure}[!t]
    \centering
    \includegraphics[width=1.0\linewidth]{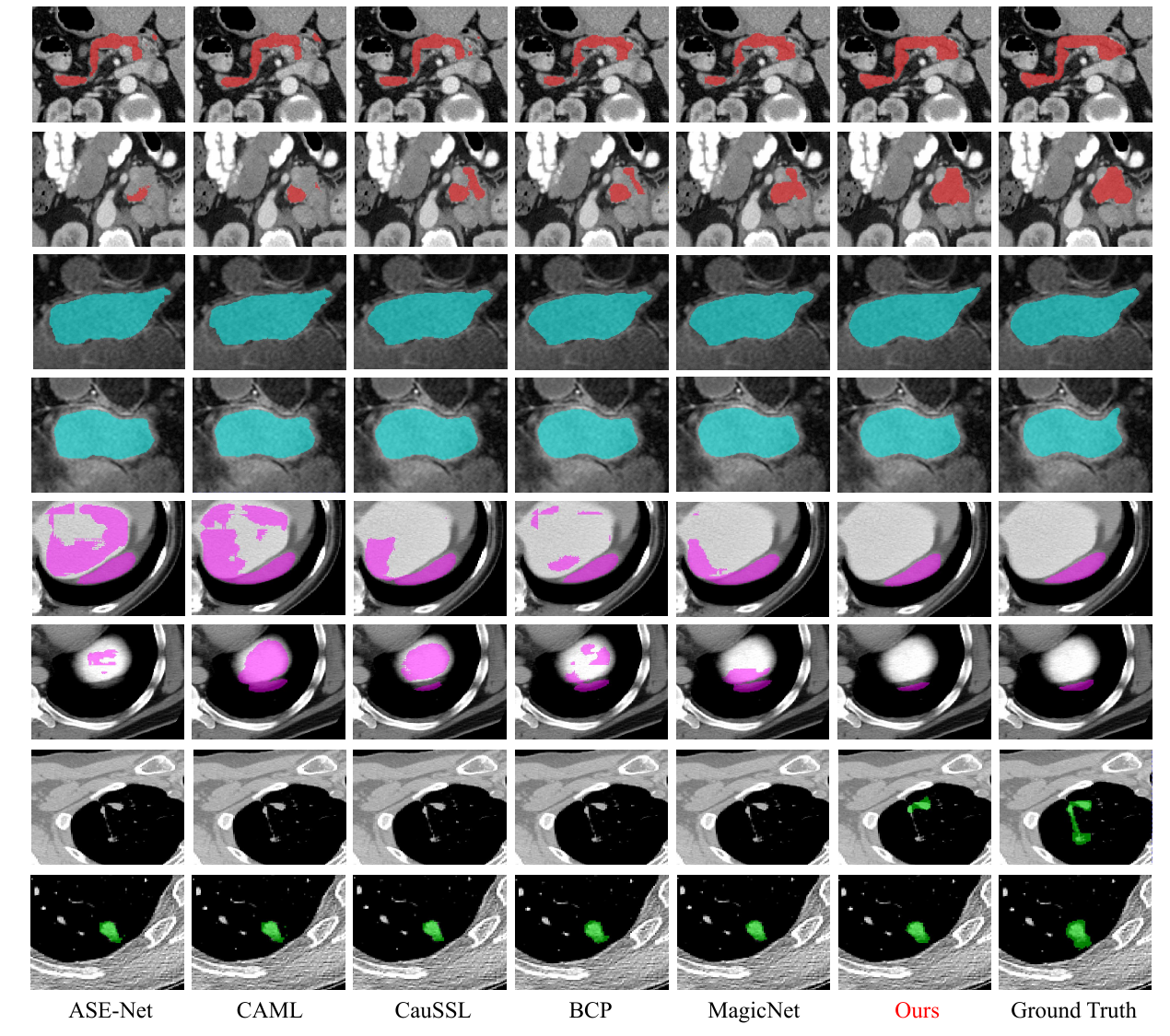}
    \caption{Segmentation results produced by different methods. Row 1-2: pancreas segmentation; Row 3-4: left atrium segmentation; Row 5-6: spleen segmentation and Row 7-8: lung tumor segmentation.}
    \label{seg_vis}
\end{figure}

{\noindent\bf Learned distribution on labeled and unlabeled data. }Distribution mismatch between labeled and unlabeled data is a commonly encountered issue in SSL, which is mainly caused by unbalanced/partial distribution learned from labeled data~\cite{zeng2023pefat, chen2023magicnet}. Fig.~\ref{kde} presents the kernel density estimation of VerSemi and BCP when training with 10\% label percentage. We can find that: (1) for Task\#2~(left atrium segmentation) with large data scale, both VerSemi and BCP show well-aligned distribution, which is mainly attributed to ample representation learned from labeled data, thus models can successfully generalize to unlabeled data and present comparable performance; (2) as for the other tasks, severe inconsistency is observed for BCP, whereas VerSemi significantly aligns the learned distribution. This demonstrates that properly learning tasks concurrently is beneficial to unlabeled data mining, since the mismatch issue between labeled and unlabeled data is largely alleviated.

{\noindent\bf Visualization of segmentation on four benchmarks.} 
Fig.~\ref{seg_vis} shows the segmentation results, it is clear to find that VerSemi can generate the most accurate mask compared to competitors. For example, for spleen segmentation (Row 5-6), other SSL methods extensively predict the background as the foreground, whereas VerSemi successfully distinguishes the region of spleen.

\section{Conclusion}

In this paper, we have presented an effective model VerSemi for semi-supervised medical image segmentation, with the new setting of integrating various tasks into a unified framework. Specifically, our VerSemi deals with different tasks in a dynamic way through the design of task prompts. A novel contrastive constraint is proposed to improve the controllability of dynamic task prompts, so as to distinguish different task information. Extensive experiments on four public datasets clearly demonstrate the effectiveness of our proposed VerSemi model, especially with limited training labels, setting new SOTA performance for semi-supervised medical image segmentation. 

{\noindent\bf Limitation and Future Work.} Since our model was trained on several available but limited datasets, the inter-dataset conflicts would unavoidably impact the training. Future work will include the study of a de-biased strategy for further investigation.


{
    \small
    \bibliographystyle{ieeenat_fullname}
    \bibliography{main}
}


\end{document}